\documentclass[runningheads]{llncs}
\usepackage{graphicx}

\usepackage{tikz}
\usepackage{comment}
\usepackage{amsmath,amssymb} 
\usepackage{color}

\usepackage{graphicx}
\usepackage{amssymb}
\usepackage{booktabs}
\usepackage{multirow}
\usepackage{pifont}
\usepackage{float}
\usepackage{caption}
\usepackage{bbm}
\usepackage{caption, subcaption, multirow, overpic, textpos}

\usepackage[accsupp]{axessibility}  

\usepackage[pagebackref,breaklinks,colorlinks]{hyperref}

\usepackage[capitalize]{cleveref}
\crefname{section}{Sec.}{Secs.}
\Crefname{section}{Section}{Sections}
\Crefname{table}{Tab.}{Tab.s}
\crefname{table}{Tab.}{Tabs.}
\usepackage[misc]{ifsym}
\usepackage{lipsum}

\newcommand{\eg}{\textit{e.g.}}%
\newcommand{\etc}{\textit{etc.}}%
\newcommand{\ie}{\textit{i.e.}}
\newcommand{\etal}{\textit{et. al.}}

\newcommand{\cmark}{\ding{51}}%
\newcommand{\xmark}{\ding{55}}%
\newcommand\blfootnote[1]{%
	\begingroup
	\renewcommand\thefootnote{}\footnote{#1}%
	\addtocounter{footnote}{-1}%
	\endgroup
}

\begin{document}
	\pagestyle{headings}
	\mainmatter
	\def\ECCVSubNumber{1677}  
	
	\title{PalGAN: Image Colorization with Palette Generative Adversarial Networks} 

	\titlerunning{PalGAN: Image Colorization with Palette Generative Adversarial Networks}
	%
	\author{Yi Wang\inst{1} \and
		Menghan Xia\inst{2} \and
		Lu Qi\inst{3} \and
		Jing Shao\inst{4} \and
		Yu Qiao\inst{1}\textsuperscript{\Letter}
	}
	\authorrunning{Y. Wang et al.}
	%
	\institute{$^1$Shanghai AI Laboratory \, $^2$Tencent AI Lab \, $^3$CUHK \, $^4$SenseTime Research\\
		\email{\{wangyi,qiaoyu\}@pjlab.org.cn \, menghanxyz@gmail.com \, luqi@cse.cuhk.edu.hk \, shaojing@senseauto.com }\\}
	\maketitle
	
	\begin{center}
		\centering
		\includegraphics[width=0.95\linewidth]{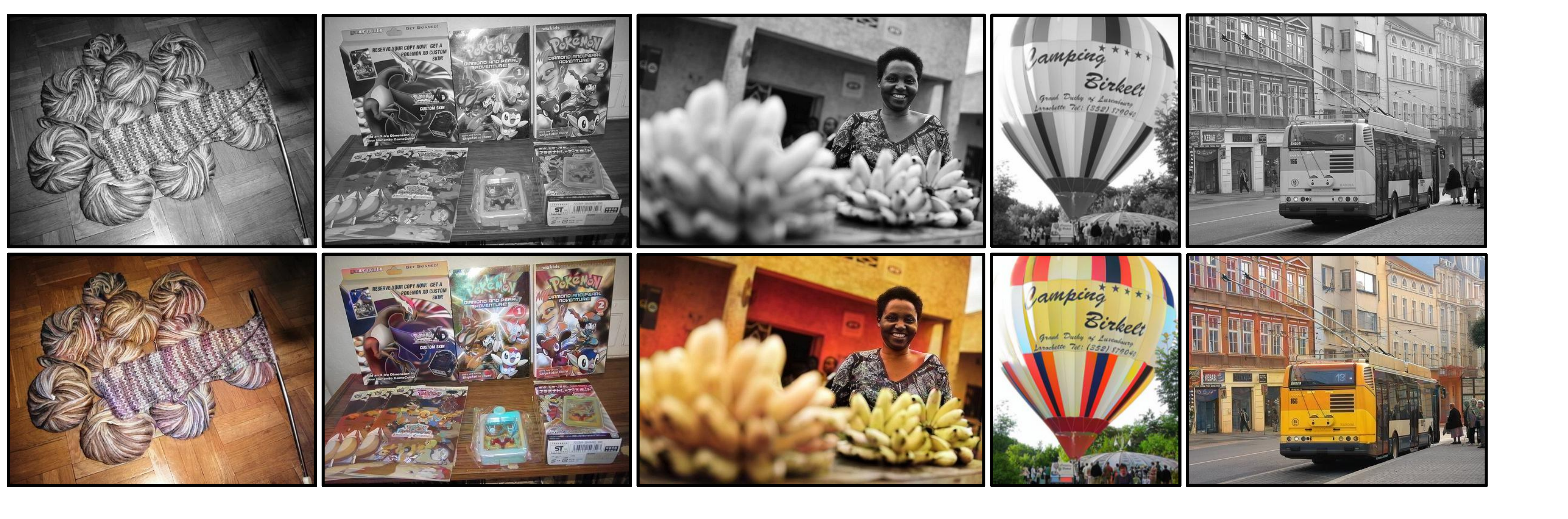}
		\vspace{-0.1in}
		\captionof{figure}{Our colorization results. $1_{st}$ row: inputs, and $2_{nd}$ row: our predictions.}
		\vspace{-0.1in}
		\label{teaser}
	\end{center}

	\begin{abstract}
		Multimodal ambiguity and color bleeding remain challenging in colorization. To tackle these problems, we propose a new GAN-based colorization approach PalGAN, integrated with palette estimation and chromatic attention. To circumvent the multimodality issue, we present a new colorization formulation that estimates a probabilistic palette from the input gray image first, then conducts color assignment conditioned on the palette through a generative model. Further, we handle color bleeding with chromatic attention. It studies color affinities by considering both semantic and intensity correlation. In extensive experiments, PalGAN outperforms state-of-the-arts in quantitative evaluation and visual comparison, delivering notable diverse, contrastive, and edge-preserving appearances. With the palette design, our method enables color transfer between images even with irrelevant contexts. \blfootnote{\Letter Corresponding author}
		\vspace{-0.1in}
	\end{abstract}
	
	\section{Introduction}
	Colorization means to predict the missing chrome information from the given gray image. It is an interesting and practical task in computer vision, widely used in legacy footage processing \cite{levin2004colorization}, color transfer \cite{afifi2021histogan,tai2005local}, and other visual editing applications \cite{bahng2018coloring,yatziv2006fast}. It is also exploited as a proxy task for self-supervised learning \cite{larsson2017colorproxy}, since predicting perceptually natural colors from the given grayscale image heavily relies on scene understanding.
	However, even the ground-truth color is available for supervision, it is still very challenging to predict pixel colors from gray images, due to the ill-posed nature that one input grayscale could correspond to multiple possible color variants.
	
	Most current methods \cite{zhang2016colorful,zhang2017real,lei2019fully,guadarrama2017pixcolor,kumar2021colorization,su2020instance,wu2021towards,iizuka2016let,bahng2018coloring} formulate colorization as a pixel-level regression task, suffering from multimodal representation more or less. With the large-scale training data and end-to-end learning models, they can learn the color distribution prior conveniently, \eg\ vegetation greenish tones, human skin colors, \etc. Anyhow, when it comes to objects with inherently color ambiguity (\eg\ human clothes, cars, and other man-made stuff), these approaches tend to predict the brownish average colors. To tackle such multi-modality, researches \cite{zhang2016colorful,zhang2017real,larsson2016learning} proposed to formulate the color prediction as pixel-level color classification, which allows multiple colors to be assigned to each pixel based on posterior probability. Unfortunately, these suffer from regional color inconsistency due to the independent pixel-wise sampling mechanism. In this regard, means of utilizing the sequential modeling~\cite{guadarrama2017pixcolor,kumar2021colorization} can only partially help the sampling issue, because the unidirectional sequential dependence of 2D flattened pixel primitives causes error accumulation and hinders the learning efficiency.
	
	Apart from the multimodal issue, color bleeding is another common issue in colorization due to inaccurate identification of semantic boundaries.
	To suppress such visual artifacts, most works~\cite{zhang2016colorful,zhang2017real,lei2019fully,su2020instance,wu2021towards,iizuka2016let,bahng2018coloring} resort to Generative adversarial networks (GAN) to encourage the generated chrome distribution to be indistinguishable from that of the real-life color images. Currently, no special algorithms or modules for deep models have been proposed to enhance the performance of this aspect, which matters the visual pleasantness considerably. 
	
	To avoid modeling the color multimodality pixel-wisely, we propose a new colorization framework PalGAN that predicts the pixel colors in a coarse-to-fine paradigm. The key idea is to first predict the global palette probability (\eg\ palette histogram) from the grayscale. It does not collapse into a single specific colorization solution but represents a certain color distribution of the potential color variants. Then, the uncertainty about the per-pixel color assignment is modeled with a generative model in the GAN framework, conditioned on the grayscale and palette histogram. Therefore, multiple colorization results could be achieved by changing the palette histogram input.
	
	To guarantee the color assignment with semantic correctness and regional consistency, we study color affinities by a proposed chromatic attention module. It explicitly aligns color affinity with both semantics and low-level characteristics. In structure, chromatic attention includes global interaction and local delineation. The former enables global context utilization for color inference by using semantic features in the attention mechanism. The latter preserves regional details by mapping the gray input to color through local affine transformation. The transformation is explicitly parameterized by the correlation between gray input and color feature.
	Experiments illustrate the effectiveness of our method. It achieves impressive visual results (Fig. \ref{teaser}) and quantitative superiority over state-of-the-art approaches over ImageNet \cite{deng2009imagenet} and COCO-Stuff \cite{caesar2018coco}. Our method also works well with the user-specified palette histogram from a reference image, which could even have no content correlation with the input grayscale. So, by nature, our method supports diverse coloring results with certain controllability. Our code and pretrained models are available in \url{https://github.com/shepnerd/PalGAN}.
	
	Generally, our contributions are three-fold: i) We propose a new colorization framework PalGAN that decomposes colorization to palette estimation and pixel-wise assignment. It circumvents the challenges of color ambiguity and regional homogeneity effectively, and supports diverse and controllable colorization by nature. ii) We explore the less-touched color affinities and propose an effective module named chromatic attention. It considers both semantic and local detail correspondence, applying such correlations to color generation.  It alleviates notable color bleeding effects. iii) Our method surpasses state-of-the-arts in perceptual quality (FID \cite{heusel2017gans} and LPIPS \cite{zhang2018perceptual}) notably. It is known that there exists a trade-off between perceptual and fidelity results in multiple low-level tasks. We argue perceptual effects matter more than fidelity as colorization aims to produce realistic colorized results rather than restore identical pixel-wise colors as the ground truth. Regardless, our method can achieve best both fidelity (PSNR and SSIM) and perceptual performance with proper tuning.
	
	\section{Related Work}
	\vspace{-0.1in}
	\subsection{Colorization}
	\vspace{-0.1in}
	\paragraph{User Guided Colorization} Some of early works \cite{charpiat2008automatic,chia2011semantic,ironi2005colorization,liu2008intrinsic,tai2005local,welsh2002transferring,kim2021deep,chang2015palette,nguyen2017group} in colorization turn to a reference image for transferring its color statistics to the given gray one. With the prevalence of deep learning, such color transfer is characterized in neural feature space for introducing semantic consistency \cite{he2018deep}. 
	These works perform decently when the reference and input share similar semantics. Its applications are limited by the reference retrieval quality, especially when handling complicated scenes.
	
	Besides of reference images, several systems require users to give sufficient local color hints (usually in scribble form) before colorizing inputs \cite{levin2004colorization,qu2006manga,yatziv2006fast,kim2021deep}. Then approaches propagate the given colors based on their local affinities. Besides, some attempts are made \cite{bahng2018coloring} to explore other modalities like languages to instruct what colors are used and how they are distributed.
	
	\vspace{-0.15in}
	\paragraph{Learning-based Colorization} This line of work \cite{zhang2016colorful,zhang2017real,deshpande2017learning,iizuka2016let,larsson2016learning,isola2017image} gives colorful images only from the gray inputs, learning a pixel-to-pixel mapping. Large-scale datasets are exploited in a self-supervised fashion, converting colorful pictures to gray ones for pair-wise training. Iizuka \etal\ \cite{iizuka2016let} utilize image-level labels for associating predicted color with global semantics, using a global-and-local convolutional neural network. Larsson \etal\ \cite{larsson2016learning} and Zhang \etal\ \cite{zhang2016colorful} introduce pixel-level color distribution matching by classification, alleviating color unbalance and multi-modal outputs. Besides, extra input hints are integrated into learning systems by simulation in \cite{zhang2017real}, providing automatic and semi-automatic ways to colorize images. Recently, transformer architectures are explored for this task considering their expressiveness on non-local modeling \cite{kumar2021colorization}. 
	
	Some work explicitly exploits additional priors from pretrained models for colorization. Su \etal\ \cite{su2020instance} study leveraging instance-level annotations (e.g., instance bounding boxes and classes) by using an off-the-shelf detector. It will make the colorization model focuses on color rendering without the need of recognizing high-level semantics. In addition to the mentioned pretrained discriminative models, pretrained generative ones are also exploitable in improving colorization performance in diversity. Wu \etal\ \cite{wu2021towards} explore to incorporate generative color prior from a pretrained BigGAN \cite{brock2018large} to help a deep model produce colored results with diversities. They design an extra encoder to project the given gray image into latent code, then estimate colorful images from BigGAN. With such primary predictions, they further refine the color results by the intermediate features in BigGAN. Afifi \etal\ \cite{afifi2021histogan} propose employing a pretrained StyleGAN \cite{karras2018style} for image recoloring, and color is controlled by histogram features. 
	
	\vspace{-0.15in}
	\subsection{GAN-based Image-to-image Translation}
	\vspace{-0.1in}
	Image-to-image translation aims to learn the transformation between the input and output image. Colorization can be formulated to this task and handled by Generative Adversarial Networks \cite{goodfellow2014generative} (GAN) based approaches \cite{isola2017image,wang2018high,park2019semantic,liu2019learning,wang2021image}. They employ an adversarial loss that learns to discriminate between real and generated images, and then minimize this loss by updating the generator to make the produced results look realistic \cite{zhu2017unpaired,li2022mat,Liu_2022_CVPR,xia2020enhance,qi2021open,xu2021conditional,wang2018inpainting,wang2019wide,wang2020vcnet}.
	
	\begin{figure*}[t]
		\centering
		\includegraphics[width=1\linewidth]{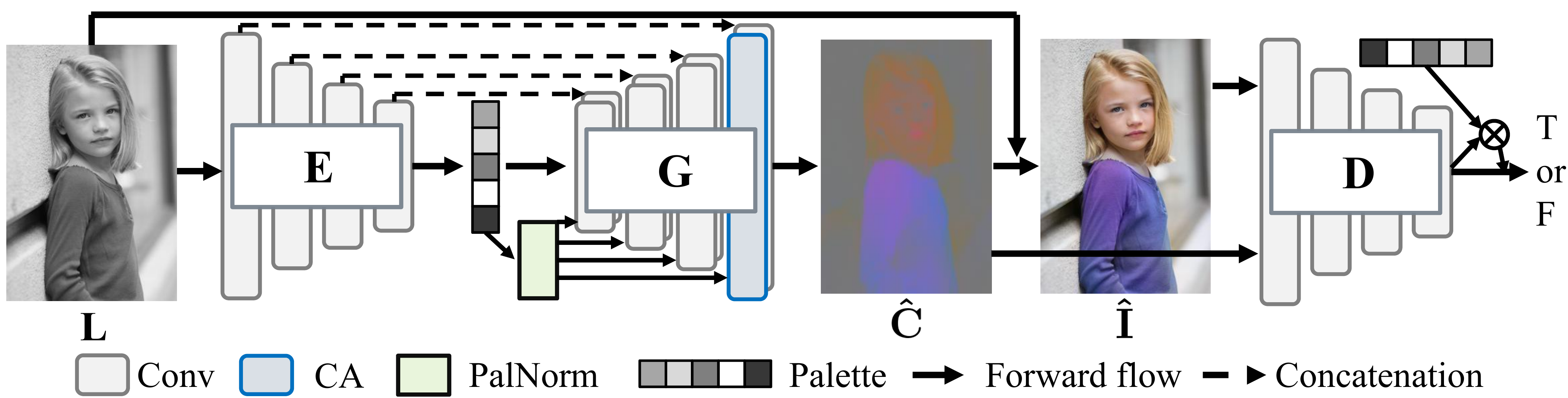}
		\vspace{-0.15in}
		\caption{Our colorization system framework.}
		\vspace{-0.15in}
		\label{fig:framework}
	\end{figure*}
	
	\section{Method}
	\label{sec:method}
	PalGAN aims to colorize grayscale images. It formulates colorization as a palette prediction and assignment problem. Compared with directly learning the pixel-to-pixel mapping from gray to color as adopted by most learning-based methods, this disentanglement fashion not only brings empirical colorization improvements (Section \ref{sec_exp}), but also enables us to manipulate global color distributions by adjusting or regularizing palettes.
	
	For PalGAN, its input is a grayscale image (\ie\ the luminance channel of color images) $\mathbf{L} \in \mathcal{R}^{h \times w \times 1}$, and the output is the estimated chromatic map $\mathbf{\hat{C}} \in \mathcal{R}^{h \times w \times 2}$ that will be used as the complementary \textit{ab} channels together with $\mathbf{L}$ in CIE \textit{Lab} color space. PalGAN consists of palette generator $\mathcal{T}_{\mathbf{E}}$, palette assignment generator $\mathcal{T}_{\mathbf{G}}$, and a color discriminator $\mathbf{D}$. In inference, only $\mathcal{T}_{\mathbf{E}}$ and $\mathcal{T}_{\mathbf{G}}$ are employed. The whole framework is given in Fig. \ref{fig:framework}.
	
	\begin{figure}[t]
		\centering
		\includegraphics[width=1.0\linewidth]{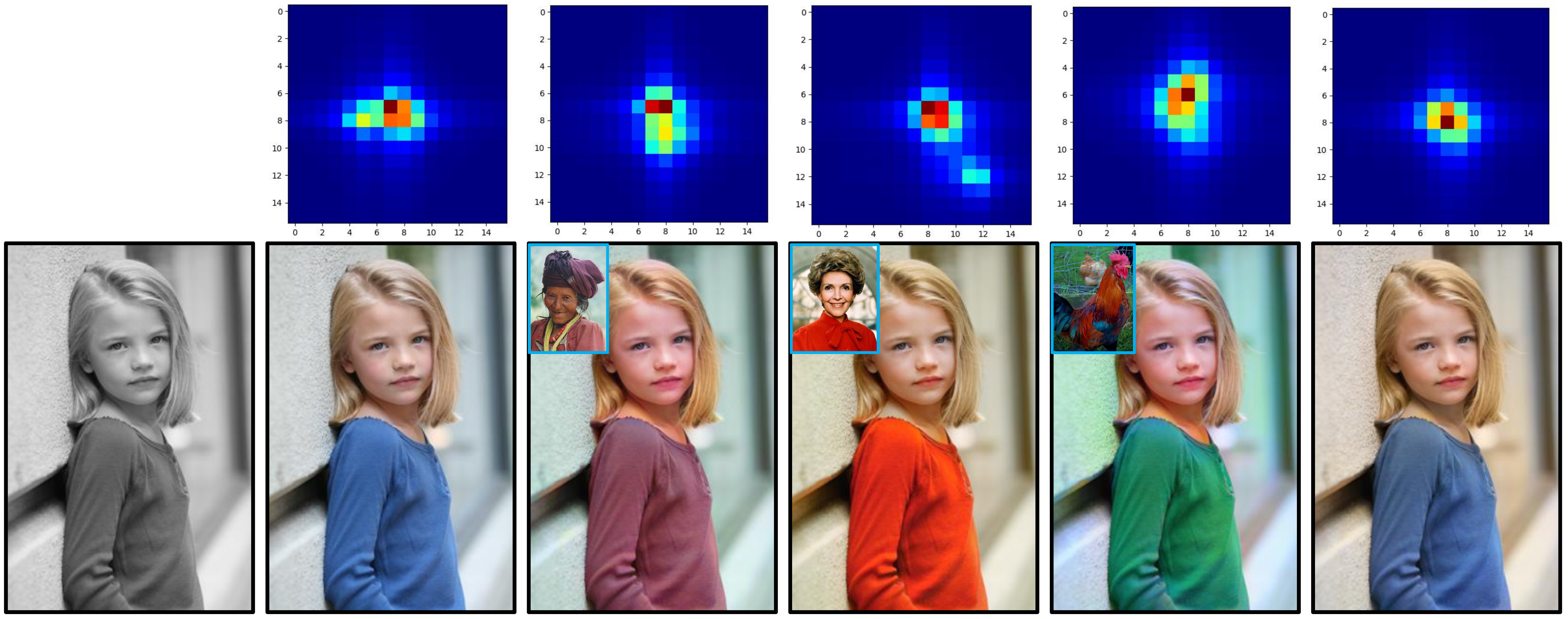}
		\small
		\begin{tabular}{cccccc} 
			\hspace{0.00\columnwidth}(a) & \hspace{0.12\columnwidth}(b) & \hspace{0.12\columnwidth}(c) & \hspace{0.12\columnwidth}(d) & \hspace{0.12\columnwidth}(e) & \hspace{0.12\columnwidth}(f)\\
		\end{tabular}
		\vspace{-0.15in}
		\caption{Visualizations of palettes (1$_\text{st}$ row, shown in jet colormap) and how they work on colorization (2$_\text{nd}$ row). (a) Input, (b) the ground truth, (c)-(e) reference-based colorization, (f) automatic colorization. }
		\vspace{-0.1in}
		\label{fig:pal}
	\end{figure}
	
	\subsection{Palette Generator}
	\label{subsec:palette_gen}
	$\mathcal{T}_{\mathbf{E}}$ estimates the global palette probabilities from the given gray image as $\mathbf{\hat{h}} = \mathcal{T}_{\mathbf{E}}(\mathbf{L})$. We employ a 2D chromatic histogram $\mathbf{\hat{h}} \in \mathcal{R}^{N_a \times N_b \times 1}$ to represent palette probabilities ($N_a$ and $N_b$ denotes bin numbers of \textit{a} and \textit{b} axes respectively), modeling the chromatic information statistics instead of learning a deterministic one. $\mathcal{T}_{\hat{P}}$ is an encoder network with several convolutional layers and a few multiple-layer perceptions (MLP), ended with a sigmoid function. The former is to extract features and the latter is to transform spatial features to a histogram (in vector form). With the explicit representation of the color palette in histogram form, we find it not only makes global color distribution more predictable, but also manipulative by introducing proper regularizations.
	
	The user-guided colorization~\cite{zhang2017real,chang2015palette,nguyen2017group} has demonstrated the effectiveness of utilizing the color histogram of a reference image for colorizing images. Compared to the existing practice~\cite{zhang2017real,chang2015palette,nguyen2017group}, we make one step further, \ie\ synthesizing a palette histogram conditioned on the input grayscale instead of taking that from a user-specified reference image. This design brings two non-trivial advantages. First, it makes our method to be a self-contained fully automatic colorization system, instead of depending on any outside guidance (i.e. a reference image) to work. Second, in general cases, the palette histogram estimated each specific grayscale may offer more accurate and instructive information for the colorization process, than that from a reference image selected in the wild. We empirically demonstrate this in Section \ref{sec_ab}. In Fig.~\ref{fig:pal}, we visualize the predicted palette histogram (f), in comparison with the ground-truth (b) and those of reference images (c~e).
	
	\vspace{-0.15in}
	\subsection{Palette Assignment Generator}
	\label{subsec:palette_assign}
\begin{figure}[t]
		\centering
		\includegraphics[width=0.750\linewidth]{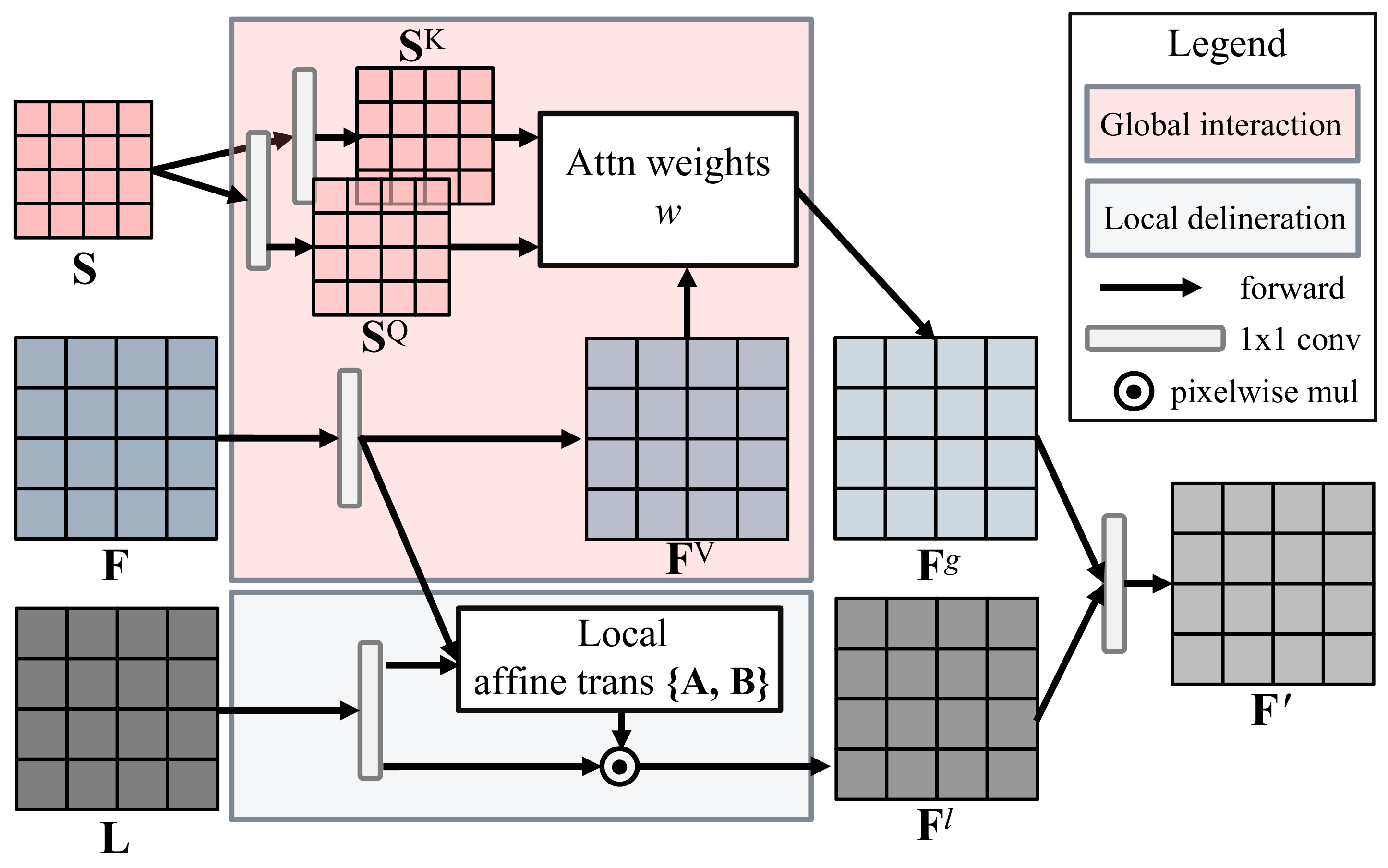}\vspace{-0.1in}
		\caption{The illustration of chromatic attention.}
		\vspace{-0.1in}
		\label{fig:hue_attn}
	\end{figure}

\begin{figure}[t]
		\centering
		\includegraphics[width=1\linewidth]{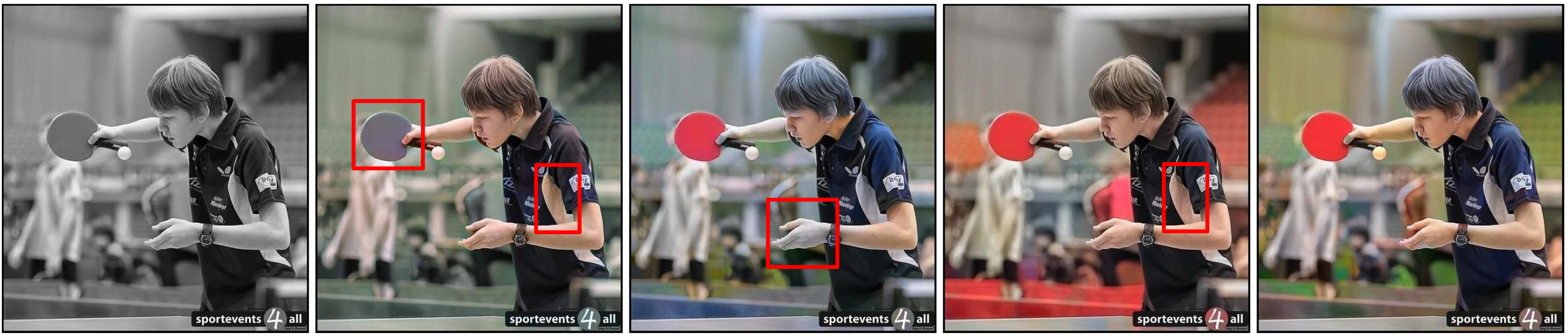}
		\small
		\begin{tabular}{ccccc} 
			\hspace{0.00\columnwidth}(a) & \hspace{0.16\columnwidth}(b) & \hspace{0.16\columnwidth}(c) & \hspace{0.16\columnwidth}(d) & \hspace{0.16\columnwidth}(e) \\
		\end{tabular}
		\vspace{-0.15in}
		\caption{Ablation studies of chromatic attention (CA). (a) input, (b) wo CA, (c) w Global, (d) w Local, (e) full CA. Please zoom in.}
		\vspace{-0.1in}
		\label{fig:ca}
	\end{figure}

	$\mathcal{T}_{\mathbf{G}}$ conducts color assignment task via conditional image generation. It produces the corresponding \textit{ab} from the gray image conditioned on palette histogram $\mathbf{\hat{h}}$ and extra latent code $z$ (sampled from a normal distribution), as $\mathbf{\hat{C}} = \mathcal{T}_{\mathbf{G}}(\mathbf{L}|\mathbf{\hat{h}},z)$. It is a convolutional generator is composed of common residual blocks used in image translation \cite{he2016deep,isola2017image}, together with our customized Palette Normalization (PN) layer and Chromatic Attention (CA) module. The palette normalization is designed to promote the conformity of the generated chromatic channels to the palette guidance $\mathbf{\hat{h}}$, which is used along with each Batch Normalization layers. Specifically, the PN layer normalizes its input feature first and then performs an affine transformation parameterized by $g(\mathbf{\hat{h}})$ (where $g(\cdot)$ is a fully-connected layer).
	Besides, we propose a chromatic attention module (Fig.~\ref{fig:hue_attn}) to explicitly align color affinity to their corresponding semantic and low-level characteristics, which mitigates potential color bleeding or semantic misunderstanding effectively. We discuss the designs below in detail, along with a visualization of the effects of its components shown in Fig.~\ref{fig:ca}.
	
	\vspace{-0.15in}
	\subsubsection{Chromatic Attention}
	The proposed Chromatic Attention (CA) module incorporates both semantic and low-level affinities into constructing color relations. These two are realized by \textit{global interaction} and \textit{local delineation} submodules (Fig.~\ref{fig:hue_attn}). Specifically, inputs to CA are a high-resolution feature map $\mathbf{F}$ (of the size $\mathcal{R}^{128 \times 128 \times 64}$ from $\mathcal{T}_{\mathbf{G}}$), high-level feature map $\mathbf{S}$, and resized gray input $\mathbf{L}$. It outputs two feature maps $\mathbf{F}^g$ and $\mathbf{F}^l$ from global interaction and local delineation respectively, and fuses them into a feature map residual, adding back to the input feature map, as:
	\begin{equation}
		\text{CA}(\mathbf{F}, \mathbf{S}, \mathbf{L}) = \mathbf{F} + \mathbf{F}' = \mathbf{F} + f(\mathbf{F}^g \oplus \mathbf{F}^l)  = \mathbf{F} + f(\text{CA}_g(\mathbf{F}|\mathbf{S}) \oplus \text{CA}_l(\mathbf{L}|\mathbf{F})),
	\end{equation}
	where $f(\cdot)$ is a nonlinear fusion operation formed by two consecutive convolutional layers, and $\oplus$ is channel-wise concatenation operation. $\text{CA}_g(\cdot)$ and $\text{CA}_l(\cdot)$ denote \textit{global interaction} and \textit{local delineation}, respectively. In this paper, we use $\mathbf{F}, \mathbf{F}' \in \mathcal{R}^{128 \times 128 \times 64}$.
	\vspace{-0.15in}
	\paragraph{Global Interaction} We reconstruct every regional feature point from the input feature map using a weighted sum of other ones, and such local weight is computed according to their semantic similarity. Formally, it is written as $\mathbf{F}_p^g = \sum_{q \in \mathbf{F}}w_{pq}\mathbf{F}^\text{V}_q$,
	where $p$ and $q$ denote a patch centering at pixel location $p$ and $q$ within $\mathbf{F}$, respectively.
	And $w_{pq}$ is calculated from the region-wise interaction in the learned high-level feature maps from the input gray images. The region-wise feature interaction is measured by the cosine similarity between the normalized regional features, as: 
	\begin{equation}
		w_{pq} = \frac{\exp(w'_{pq})}{\sum_{k \in \mathbf{S}}\exp(w'_{pk})} \quad \text{where} \quad w'_{pq} = \frac{\mathbf{S}^\text{K}_p \cdot \mathbf{S}^\text{Q}_q}{|\mathbf{S}^\text{K}_p| |\mathbf{S}^\text{Q}_q|},
	\end{equation}
	where $\mathbf{S}$ denote high-level feature map, extracted from intermediate representation of the encoder $\mathcal{T}_{\hat{P}}$. $\mathbf{S}^\text{K}$ and $\mathbf{S}^\text{Q}$ denote two translated feature maps from $\mathbf{S}$ using convolution.
	
	\vspace{-0.15in}
	\paragraph{Local Delineation} Though color changes in texture and edges are delicate, overlooking these subtle variances leads to notable visual degradation. To preserve these details, we design a local delineation module to complement global interaction. We adopt the assumption that local color affinity is linearly correlated with its corresponding intensity \cite{zomet2002multi,torralba2002properties}. We propose to learn such local relationship in the guided filter manner \cite{he2012guided,wu2018fast}, which preserves edges from the guidance well. Our given local preserving module computes a learnable local affine transformation $\{\mathbf{A} \in \mathcal{R}^{128 \times 128 \times 64}, \mathbf{B}  \in \mathcal{R}^{128 \times 128 \times 64}\}$ to map the gray image $\mathbf{L} \in \mathcal{R}^{256 \times 256 \times 1}$ to its corresponding \textit{ab} feature map, as:
	\begin{equation}
		\mathbf{F}^l = \mathbf{A} \odot \mathbf{L}\downarrow + \mathbf{B},
	\end{equation}
	where $\odot$ is the element-wise multiplication operator and $\downarrow$ is downsampling one to ensure the spatial size of $\mathbf{L}$ is the same as $\mathbf{F}^{l}$. $\{\mathbf{A}, \mathbf{B}\}$ are parameterized by the a learnable local correlation between $\mathbf{L}$ and $\mathbf{F}$, as:
	\begin{equation}
		\mathbf{A} = \Psi(\frac{\text{cov}(\mathbf{F}, \mathbf{L})}{\text{var}(\mathbf{L})+\epsilon}), \;
		\mathbf{B} = \overline{\mathbf{F}} - \mathbf{A} \odot \overline{\mathbf{L}}
	\end{equation}
	where $\Psi$ is a learnable transformation parameterized by a small convolutional net, $\text{cov}(\cdot,\cdot)$ computes the local covariance between two feature maps (within a fixed window size) while $\text{var}()$ computes the local variance of the given feature map. $\overline{\mathbf{F}}$ and $\overline{\mathbf{L}}$ denote the smoothed versions of $\mathbf{F}$ and $\mathbf{L}$ by a mean filter, respectively. $\epsilon$ is a small positive number for computational stability.
	
	\vspace{-0.15in}
	\subsubsection{Palette Optimization}
	To further ensure the proposed palette assignment generator is responsive to the given palette, we minimize the discrepancy between the palette extracted from the predicted chromatic channels and that from the corresponding ground truth.
	However, common histograms from images are non-differentiable due to the hard thresholds. Follow the practice of~\cite{afifi2021histogan}, we regard the palette histogram as a joint distribution over \textit{a} and \textit{b}, represented by a weighted sum of kernels.
	Formally, the color histogram is written as:
	\begin{equation} \label{eq_palette}
		\mathbf{h}(a, b) = \frac{1}{Z}\sum_{x}k(\mathbf{C}_a(x), \mathbf{C}_b(x), a, b),
	\end{equation} 
	where $\mathbf{C}_a(x)$ and $\mathbf{C}_b(x)$ denote the values of pixel $x$ in \textit{a} and \textit{b} channels, respectively. $k$ is the used kernel function to measure the difference between $(\mathbf{C}_a(x), \mathbf{C}_b(x))$ and a given $(a, b)$, $Z$ is a normalization factor. In this paper, we adopt inverse-quadratic kernel \cite{afifi2021histogan}, which is:
	\begin{equation} \small
		k(\mathbf{C}_a(x), \mathbf{C}_b(x), a, b)= \prod_{i \in \{a, b\}} (1+(\frac{|\mathbf{C}_i(x)-i|}{\sigma})^2)^{-1},
	\end{equation}
	where $\sigma$ controls the smoothness of adjacent bins. We find $\sigma=0.1$ works best.
	
	\vspace{-0.15in}
	\paragraph{Regularization}To diversify the predicted colors, we introduce palette regularization, combating against the dull colors brought by imbalanced color distribution. On one hand, we employ \textit{ab} histogram in probabilistic palette form to measure the color distribution in the predicted color map and ground truth. Minimizing their discrepancies explicitly considers different color ratios, avoiding converging to a few dominant ones. On the other hand, we diversify the produced colors by increasing the possibility of rare colors (statistically in training samples). We exploit the entropy of the probabilistic palette to control such diversity. Formally, the entropy of $\mathbf{\hat{h}}$ is
	$E(\mathbf{\hat{h}})=-\sum_{i=1}^{|\mathbf{\hat{h}}|}\mathbf{\hat{h}}_{i}\log\mathbf{\hat{h}}_{i}$.
	To improve the color diversity in $\mathbf{\hat{h}}$, we can maximize $E(\mathbf{\hat{h}})$.
	
	\vspace{-0.15in}
	\subsection{Color Discriminator}
	We give a color discriminator utilizing the palette, improving the result from the adversarial training. We incorporate the palette into the discriminator in a condition projection manner \cite{miyato2018cgans}. We employ convolutional discriminator $\mathbf{D}$, converting the input (the concatenation between the \textit{ab} image and its converted RGB one) into a 1D feature embedding $\mathbf{g \in \mathbb{R}^{256 \times 1}}$. Then such feature is fused with the palette by the inner product. The likelihood of the realness of the input is given as:
	\begin{equation}
		p(\mathbf{C} \oplus \mathbf{I}) = (\mathbf{W}\mathbf{g}) ^{\text{T}} \mathbf{h},
	\end{equation}
	where $\mathbf{W} \in \mathbb{R}^{n^2 \times 256}$ is a learnable linear projection, and $\mathbf{I} \in \mathbb{R}^{h \times w \times 3}$ is the converted \textit{rgb} version of $\mathbf{C}$ and $\mathbf{L}$. 
	
	\vspace{-0.15in}
	\subsection{Learning Objective}
	Palette estimation and assignment are trained with different optimization targets. For palette estimation, it is learned concerning palette reconstruction and regularization as:
	\begin{equation}
		\mathcal{L}_\mathbf{E} = \underbrace{\lambda_\text{rec1}|\mathbf{h}-\mathbf{\hat{h}}|_1}_{\text{reconstruction}} - \underbrace{\lambda_\text{rg}E(\mathbf{\hat{h}})}_{\text{regularization}},
	\end{equation}
	where $\lambda_\text{rec1}$ and $\lambda_\text{rg}$ balance the influences of different terms, set to 5.0 and 1.0, respectively.
	
	The optimization target for palette assignment is formed by pixel-level regression, palette reconstruction, and adversarial training, as:
	\begin{equation} \small
		\mathcal{L}_\mathbf{G} = \underbrace{\lambda_\text{reg}|\mathbf{C}-\mathbf{\hat{C}}|_1}_{\text{regression}} + \underbrace{\lambda_\text{rec2} |\mathbf{h}-\mathbf{\tilde{h}}|_1}_{\text{reconstruction}} + \underbrace{\lambda_\text{adv} \mathcal{L}_\text{adv}}_{\text{adversarial}},
	\end{equation}
	where $\mathbf{\tilde{h}}$ are extracted from $\mathbf{\hat{C}}$ using Eqn. \ref{eq_palette}. $\lambda_\text{reg}$, $\lambda_\text{rec2}$, and $\lambda_\text{adv}$ are set to 5.0, 1.0, 1.0, respectively.
	
	For the used adversarial loss, we employ hinge loss version. Its training target of generator is
	\begin{equation}
		\mathcal{L}_{adv} =  -\text{E}_{\mathbf{L} \sim \mathbb{P}_\mathbf{L}}\mathbf{D}(\mathbf{\hat{C}} \oplus \mathbf{\hat{I}}),
	\end{equation}
	where $\mathbf{\hat{C}}=\mathcal{T}_{\mathbf{G}}(\mathbf{L}|\mathcal{T}_{\mathbf{E}}(\mathbf{L}))$, $\mathbf{\hat{I}}$ is a converted \textit{rgb} version from $\mathbf{\hat{C}}$ and $\mathbf{L}$, and $\mathbb{P}_\mathbf{L}$ denotes the gray-scale image distribution. The optimization goal for the discriminator is
	\begin{equation}
		\mathcal{L}_{adv}^{\mathbf{D}} = \text{E}_{\mathbf{I} \sim \mathbb{P}_\mathbf{I}}[\text{max}(0, 1-\mathbf{D}(\mathbf{C} \oplus \mathbf{I}))]+ \text{E}_{\mathbf{L} \sim \mathbb{P}_\mathbf{L}}[\text{max}(0, 1+\mathbf{D}(\mathbf{\hat{C}} \oplus \mathbf{\hat{I}})].
	\end{equation}
	where $\mathbb{P}_\mathbf{I}$ denotes the \textit{rgb} image distribution, and $\mathbf{C}$ is converted from $\mathbf{I}$.
	
	\vspace{-0.15in}
	\paragraph{Training}
	We jointly train palette generator $\mathcal{T}_{\mathbf{E}}$ and palette assignment generator $\mathcal{T}_{\mathbf{G}}$ in an progressive fashion. Specifically, for the inputs $\{\mathbf{L}_i\}$ to $\mathcal{T}_{\mathbf{E}}$, the corresponding inputs to $\mathcal{T}_{\mathbf{G}}$ are $\{\mathbbm{1}(p_{\mathbf{h}} > 0.8) \mathbf{h}_i+(1-\mathbbm{1}(p_{\mathbf{h}} > 0.8))\hat{\mathbf{h}}_i\}$, where $\mathbbm{1}$ is an indicator function of value $1$ if its condition holds true, and $0$, otherwise. $p_{\mathbf{h}}$ is sampled from a uniform distribution $\mathcal{U}[\tau, 1]$. We start training with $\tau=1$, then linearly decrease it to $0$ when approaching the end of learning.
	
	\begin{table}[t]
		\centering
		\caption{Quantitative results on the validation sets from different methods.}
		\label{tb_me_evaluation1}
		\small
		\setlength{\tabcolsep}{0.1mm}{
			\resizebox{\textwidth}{22mm}{
				\begin{tabular}{c|ccccccccccccc}
					\toprule
					\multirow{2}*{Method} & \multicolumn{4}{c}{ImageNet (ctest10k)} & \multicolumn{4}{c}{ImageNet (val50k)} & \multicolumn{4}{c}{COCO-Stuff}  \\
					~ & PSNR $\uparrow$ & SSIM $\uparrow$ & LPIPS $\downarrow$ & FID $\downarrow$ & PSNR $\uparrow$ & SSIM $\uparrow$ & LPIPS $\downarrow$ & FID $\downarrow$ & PSNR $\uparrow$ & SSIM $\uparrow$ & LPIPS $\downarrow$ & FID $\downarrow$ \\
					\midrule
					CIColor\ \cite{zhang2016colorful} & 22.30 & 0.902 & 0.221 & 12.20 & 22.26 & 0.902 & 0.221 & 9.39 & 21.84 & 0.895 & 0.234 & 22.32\\
					UGColor \cite{zhang2017real} & 24.26 & 0.918 & 0.174 & 7.49 & 24.26 & 0.919 & 0.173 & 4.60 & 24.34 & 0.924 & 0.165 & 14.74\\
					Lei \etal\ \cite{lei2019fully} & 24.52 & 0.917 & 0.202 & 12.60 & 24.03 & 0.918 & 0.189 & 6.35 & 24.59 & 0.922 & 0.191 & 23.10\\
					Deoldify \cite{antic} & 23.54 & 0.914 & 0.187 & 5.78 & 22.97 & 0.911 & 0.185 & 3.87 & 23.98 & 0.939 & 0.161 & 12.75\\
					ColTrans \cite{kumar2021colorization} &21.81  &0.892  &0.218  &6.37  &22.12  &0.894  &0.216 &3.81 & 22.11  &0.898  &0.210  &11.65\\
					Ours$^1$ & 24.19 & 0.917 & \textbf{0.161} & \textbf{4.60} & 24.25 & 0.917 & \textbf{0.161} & \textbf{2.78} & 24.56 & 0.924 & \textbf{0.148} & \textbf{7.70}\\
					Ours$^2$ & \textbf{24.66} & \textbf{0.920} & 0.170 & 5.24 & \textbf{24.54} & \textbf{0.920} & 0.168 & 3.62 & \textbf{24.72} & \textbf{0.944} & 0.156 & 8.93\\
					\midrule
					InstColor* \cite{su2020instance} & 23.03 & 0.909 & 0.191 & 7.35 & 23.06 & 0.910 & 0.190 & 4.94 & 22.35 & 0.838 & 0.238 & 12.24\\
					GPColor* \cite{wu2021towards} & 21.66 & 0.871 & 0.230 & 5.46 & 21.81 & 0.880 & 0.230 & 3.62 & N/A & N/A & N/A & N/A\\
					\midrule
					
					Ours* & 27.75 & 0.932 & 0.110 & 4.20 & 27.53 & 0.913 & 0.118 & 2.42 & 28.28 & 0.936 & 0.105 & 7.21\\
					
					\bottomrule
				\end{tabular}
			}
		}
		\vspace{-0.15in}
	\end{table}
	
	\section{Experiments} \label{sec_exp}
	
	We evaluate our method along with existing representative works on ImageNet \cite{deng2009imagenet} and COCO-Stuff \cite{caesar2018coco}. On ImageNet we take two evaluation protocols. One is to evaluate all methods on a selective subset \textit{ctest10k} (with 10K pictures) of its validate data (with 50K pictures) following the protocols in \cite{larsson2016learning}. Another is to run on the full validation set, same as in \cite{wu2021towards}. For COCO-Stuff, we test all methods on its 5K validating images.
	\vspace{-0.15in}
	\subsection{Implementation}
	We employ spectral normalization \cite{miyato2018spectral} on the whole model and a two time-scale update rule in training ($lr$ for the generator and discriminator are $1e-4$ and $4e-4$, respectively) to stabilize learning. Adam \cite{kingma2014adam} optimizer with $\beta_1=0$ and $\beta_2=0.9$ is used. For the applied batch normalization, we take the sync version. We train our method on the training set of ImageNet with 40 epochs with 8 TiTAN 2080ti using batch size 64. Images in training are randomly cropped in a fixed size (256 $\times$ 256) from the resized ones with aspect ratio unchanged. In testing, we resize images into 256 $\times$ 256 ones and do evaluations.
	
	\vspace{-0.15in}
	\paragraph{Baselines} We focus on the recent learning-based colorization methods for comparisons. Deoldify\cite{antic}, CIColor \cite{zhang2016colorful}, UGColor \cite{zhang2017real}, Video Colorization \cite{lei2019fully}, InstColor \cite{su2020instance}, ColTrans \cite{kumar2021colorization}, and GPColor \cite{wu2021towards} are employed for comparisons. Note 
	InstColor is learned with a pretrained object detection model (requiring both labels and bounding boxes), and GPColor exploits a pretrained (on ImageNet with labels) BigGAN. Other approaches including ours are only trained with paired gray-colorful images. For UGColor, we use its fully automatic version where no color hints are used. We use their released model for testing.
	
	\vspace{-0.15in}
	\paragraph{Metrics} We employ pixel-wise similarity measures PSNR, SSIM, image-level perceptual metric LPIPS \cite{zhang2018perceptual}, and Fr$\acute{\text{e}}$chet Inception Distance (FID) \cite{heusel2017gans} to quantitatively evaluate colorization results. LPIPS and FID are more consistent with human evaluations compared with PSNR and SSIM.
	
	\vspace{-0.15in}
	\subsection{Quantitative Evaluations}
	Compared with other methods, our proposed PalGAN (ours$^1$ in Tab. \ref{tb_me_evaluation1}) gives the best perceptual scores (FID \& LPIPS) both on ImageNet (FID: 4.60 and 2.78, LPIPS: 0.161 and 0.161 from ctest10K and val50K, respectively) and COCO-Stuff (FID: 7.70, LPIPS: 0.148) without exploiting any annotations or hints, which outperforms other methods. It validates the superiority in realness and diversity of our results. We also achieve competitive fidelity scores (PSNR \& SSIM) among all. It shows the well-behaved color restoration ability of PalGAN. If given the ground truth palette, our method (ours* in Tab. \ref{tb_me_evaluation1}) can deliver impressive fidelity performance as well as a generative one. It shows the upper bound performance of our method for reference. For methods in Tab. \ref{tb_me_evaluation1} denoted with *, they employ external priors \eg\ annotations. 
	
	Considering the trade-off between fidelity and perceptual results, we can get the best of both worlds on all benchmarks compared with others (ours$^2$ in Tab. \ref{tb_me_evaluation1}) with proper training setting ($\lambda_{adv}=0.1$ and other regularization coefficients remain still).
	
	\begin{figure*}[t]
		\begin{center}
			\includegraphics[width=1\linewidth]{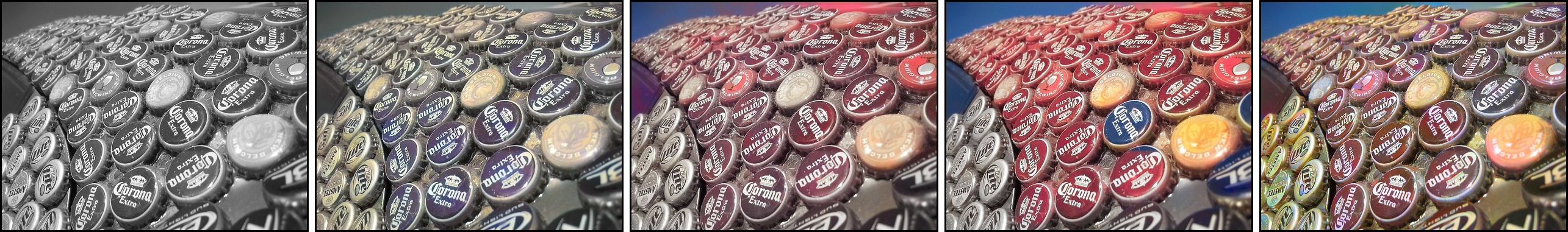} 
			\includegraphics[width=1\linewidth]{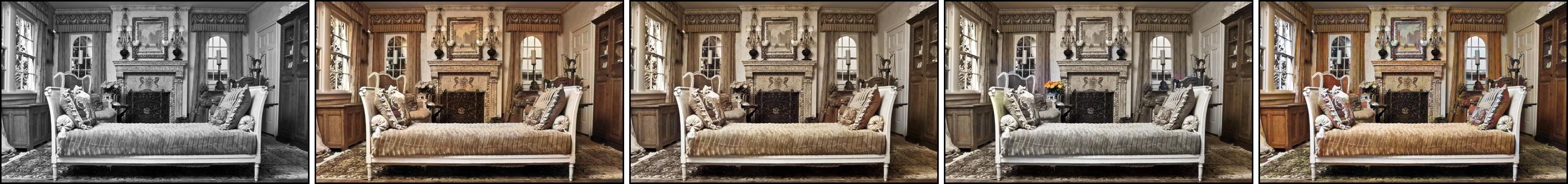} 
			\includegraphics[width=1\linewidth]{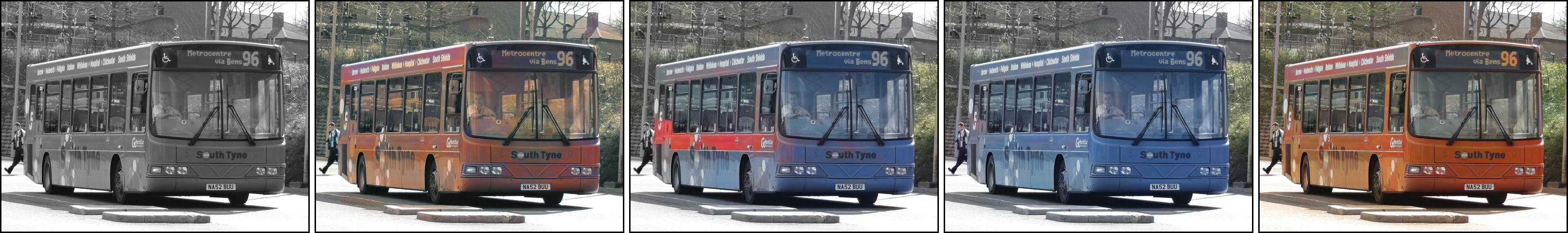}
			\includegraphics[width=1\linewidth]{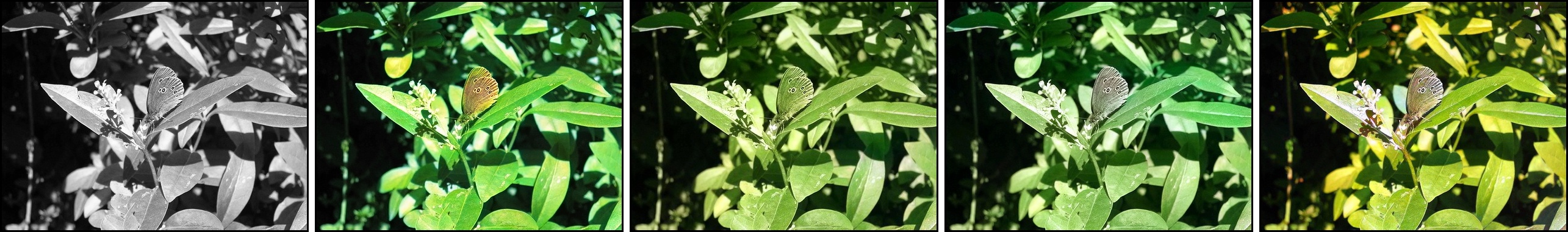} 
		\end{center}
		\vspace{-0.15in}
		\small
		\begin{tabular}{ccccc}
			\hspace{0.06\columnwidth}(1) Input & \hspace{0.06\columnwidth}(2) Deoldify & \hspace{0.04\columnwidth}(3) UGColor & \hspace{0.04\columnwidth}(4) InstColor & \hspace{0.06\columnwidth}(5) Ours \\
		\end{tabular}
		\vspace{-0.1in}
		\caption{Visual comparison on ImageNet and COCO-Stuff.}
		\vspace{-0.1in}
		\label{fig_exp1}
	\end{figure*}
	
	\begin{table}[t]
		\centering
		\caption{User study. Each entry gives the percentage of cases where colorization results are favored compared with GT.}
		\label{tb_user_studies}
		\small
		\setlength{\tabcolsep}{1mm}{
			\begin{tabular}{c|cccccc}
				\hline
				Method & Ours & Coltrans & GPCol & InstCol & Deoldify & UGCol\\
				\hline
				Rate & \textbf{47.20}\% & 41.50\% & 39.25\% & 37.50\% & 41.13\% & 42.50\%\\
				\hline
			\end{tabular}
		}
		\vspace{-0.15in}
	\end{table}
	
	\subsection{Qualitative Evaluations}
	As shown in Fig. \ref{fig_exp1}, our colorization results give natural, diverse, and fine chrome predictions considering both semantic correspondence and local gradient change. It suffers less from the common color bleeding compared with other methods, owing to chromatic attention. More results are given in Supp.
	\vspace{-0.15in}
	\paragraph{User Studies} Tab. \ref{tb_user_studies} gives human evaluations on our methods with the compared ones. Following the protocol in \cite{zhang2016colorful,kumar2021colorization}, we conduct a colorization Turning test. Specifically, the ground truth color image and its corresponding colorization result (from ours or other methods) are given to 20 participants in random order. These participants need to determine which one is more realistic than the other for no more than 2 seconds. There are 40 colorization predictions from each method, randomly chosen from ImageNet ctest10k. Tab. \ref{tb_user_studies} presents that our method beats the competitors with a large margin.
	
	\begin{figure}[t]
		\centering
		\includegraphics[width=1.0\linewidth]{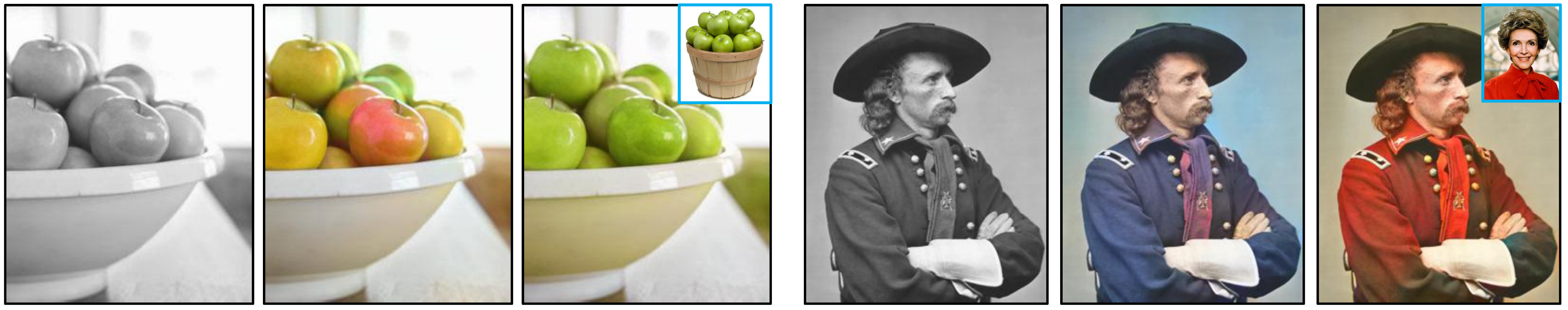}
		\small
		\begin{tabular}{cccccc} 
			\hspace{0.00\columnwidth}(1) & \hspace{0.12\columnwidth}(2) & \hspace{0.12\columnwidth}(3) & \hspace{0.14\columnwidth}(1) & \hspace{0.12\columnwidth}(2) & \hspace{0.12\columnwidth}(3)\\
		\end{tabular}
		\vspace{-0.1in}
		\caption{Our method on legacy images. (1) Inputs, (2) our automatic results, (3) our reference-based results.}\vspace{-0.1in}
		\label{fig:legacy}
	\end{figure}
	
	\vspace{-0.15in}
	\paragraph{Colorization of Legacy Photos}
	Though our model is trained in a self-supervised manner using synthetic data, it generalizes well on real-world black-and-white legacy pictures (from \cite{he2018deep}), as given in Fig. \ref{fig:legacy}. Color boundaries and consistency are well handled in these cases, working well on the object and portrait.
	
	\begin{figure*}[!t]
		\begin{center}
			\includegraphics[width=1\linewidth]{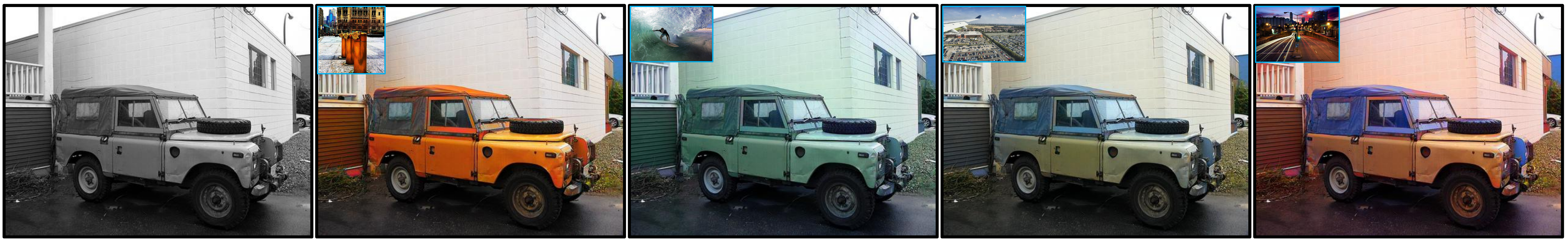}
		\end{center}
		\vspace{-0.15in}
		\small
		\begin{tabular}{cc}
			\hspace{0.08\columnwidth}Input & \hspace{0.24\columnwidth}Reference-based Colorization  \\
		\end{tabular}
		\vspace{-0.1in}
		\caption{Our reference-based colorization.}
		\label{fig_ref}
		\vspace{-0.1in}
	\end{figure*}
	
	\vspace{-0.15in}
	\paragraph{Reference-based Colorization}
	With the intermediate palette, our approach can conduct reference-based (or example-based) colorization by feeding it with the palette from the reference color image, as given in Fig. \ref{fig:legacy} and \ref{fig_ref}. Note even using palettes from an image without semantic correlations with the input (Fig. \ref{fig_ref}), PalGAN still well tunes the given color distribution according to the semantics of the given image, keeping color regionally consistent. Note the car appearances in Fig. \ref{fig_ref} present impressive diversity and realness.
	
	\begin{figure}[t]
		\centering
		\includegraphics[width=1.0\linewidth]{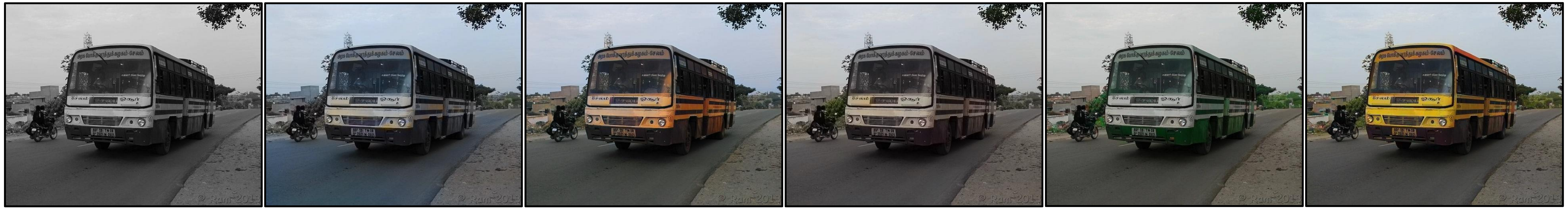}
		\small
		\begin{tabular}{cccccc} 
			\hspace{0.00\columnwidth}(a) & \hspace{0.12\columnwidth}(b) & \hspace{0.12\columnwidth}(c) & \hspace{0.12\columnwidth}(d) & \hspace{0.12\columnwidth}(e) & \hspace{0.12\columnwidth}(f)\\
		\end{tabular}
		\vspace{-0.15in}
		\caption{Ablation studies of model structures. (a) input, (b) AE, (c) VAE, (d) PalGAN w PatchD, (e) PalGAN wo $E(\mathbf{\hat{h}})$. (f) full PalGAN.}
		\vspace{-0.1in}
		\label{fig:structure}
	\end{figure}
	
	\vspace{-0.15in}
	\subsection{Ablation Studies} \label{sec_ab}
	Our key designs are ablated on COCO-Stuff as follows. 
	
	\vspace{-0.15in}
	\paragraph{Palette Prediction and Assignment} We validate the effectiveness of our colorization formulation with the proposed model structure compared with a naive autoencoder (AE) and variational one (VAE). Specifically, AE shares the same computational units with PalGAN, except it generates feature maps instead of the palette from its encoder, and utilizes common BN instead of PalNorm in its decoder. VAE is almost the same as PalGAN but it changes the intermediate product palette into a latent vector constrained by Normal distribution. The optimization of AE and VAE is nearly the same as ours except they do not have the palette reconstruction and regularization term, and VAE employs one more term for regularizing the intermediate latent code. 
	
	In Tab. \ref{tb_ab_structure}, we find PalGAN gets significant improvements on FID compared with AE and VAE, while its fidelity performance (PSNR and SSIM) is inferior to AE. It suggests intermediate latent code (in PalGAN and VAE) performs better at color generation than feature maps (in AE), and feature maps excel at fidelity restoration. It validates the effectiveness of our formulation and method on the usage of palette considering its generative performance. Moreover, Fig. \ref{fig:structure} illustrates visual differences between varied structures in one example. A high fidelity score of AE does not guarantee the realism of its result.
	
	The effectiveness of the predicted palette is studied. We use palettes from random reference images to simulate failed palette estimations in our method, and give the corresponding evaluation in Tab. \ref{tb_ab_structure} (PalGAN w rand ref). It shows the dramatic fidelity and generative performance drop, meaning our palette generator can learn effective chrome distribution for colorization. This is also supported by the visualizations of palettes and their corresponding images in Fig.~\ref{fig:pal}.
	
	\begin{table}[t]
		\centering
		\caption{Quantitative results on COCO-Stuff using different structures.}
		\label{tb_ab_structure}
		\small
		\setlength{\tabcolsep}{1mm}{
			\begin{tabular}{c|cccc}
				\toprule
				Structure & PSNR $\uparrow$ &
				SSIM $\uparrow$ &
				LPIPS $\downarrow$ &
				FID $\downarrow$ \\
				\midrule
				AE & \textbf{25.89} & \textbf{0.928} & \textbf{0.146} & 14.15\\
				VAE & 23.21 & 0.905 & 0.179 & 11.76\\
				UGC w CA & 24.52 & 0.923 & 0.162 & 11.38\\
				PalGAN w rand ref& 20.88 & 0.883 & 0.240 & 9.64\\
				PalGAN w SA & 22.68 & 0.892 & 0.175 & 9.02\\
				PalGAN w PatchD \cite{isola2017image,wang2018high} & 23.07 & 0.895 & 0.183 & 8.44\\
				PalGAN w BN & 22.36 & 0.895 & 0.209 & 9.97\\
				PalGAN w SPADE \cite{park2019semantic} & 24.06 & 0.916 & 0.167 & 7.90\\
				PalGAN wo $E(\mathbf{\hat{h}})$ & 24.58 & 0.924 & 0.149 & 8.17\\
				PalGAN & 24.56 & 0.924 & 0.148 & \textbf{7.70}\\
				\bottomrule
			\end{tabular}
		}
		\vspace{-0.1in}
	\end{table}
	
	\vspace{-0.15in}
	\paragraph{Chromatic Attention} We explore how the proposed chromatic attention affects colorization, given in Tab. \ref{tb_ab_structure} and \ref{tb_ab_hue_attention}. Compared with naive self-attention (PalGAN w SA in Tab. \ref{tb_ab_structure}, and SA is applied on the high-level feature maps $\mathbf{S}$), our chromatic attention enhances both generative and fidelity performance notably. In Tab. \ref{tb_ab_hue_attention}, with global interaction in chromatic attention, the generative performance will be improved non-trivially on FID ($9.90 \rightarrow 8.34$). It is consistent with the observations in prior image generation works \cite{zhang2018self,brock2018large,wang2020attentive,wang2021image} that employing attention will boost generation results. For the local delineation, it focuses on pixel-level restoration, giving notable fidelity increase on PSNR ($21.93 \rightarrow 24.52$) and SSIM ($0.902 \rightarrow 0.924$). Generally, CA achieves the best of both worlds as it enhances both pixel- and perceptual-level performance. Moreover, we give visual comparison of the ablation study on the chromatic attention in Fig. \ref{fig:ca}.
	
	Note CA is a generic parametric module. It can be applied to previous methods \eg\ UGC \cite{zhang2017real}, and it can further improve the corresponding quantitative results ($24.34 \rightarrow 24.51$, $0.924 \rightarrow 0.925$, $0.165 \rightarrow 0.162$, and $14.74 \rightarrow 11.38$ on PSNR, SSIM, LPIPS, and FID, respectively).
	
	\begin{table*}[t]
		\vspace{-.2em}
		\centering
		{
			\centering
			\begin{minipage}{0.42\linewidth}{\begin{center}
						\caption{Quantitative results on COCO-Stuff by ablating chromatic attention.}
						\label{tb_ab_hue_attention}
						\begin{tabular}{cc|cccc}
							\toprule
							G & L & {PSNR $\uparrow$} &
							{SSIM $\uparrow$} &
							{LPIPS $\downarrow$} &
							{FID $\downarrow$} \\
							\midrule
							\xmark & \xmark & 21.93 & 0.902 & 0.203 & 9.90\\
							\xmark & \cmark & 24.52 & \textbf{0.924} & \textbf{0.146} & 9.97\\
							\cmark & \xmark & 23.32 & 0.907 & 0.174 & 8.34\\
							\cmark & \cmark & \textbf{24.56} & \textbf{0.924} & 0.148 & \textbf{7.70}\\
							\bottomrule
						\end{tabular}
			\end{center}}\end{minipage}
		}
		\hspace{2em}
		{
			\begin{minipage}{0.46\linewidth}
				{\begin{center}
						\caption{Quantitative results on COCO-Stuff about palette with different bins.}
						\label{tb_ab_bins}
						\begin{tabular}{c|ccccc}
							\toprule
							\#Bins & 16 & 64 & 256&
							576 &
							1024 \\
							\midrule
							PSNR $\uparrow$ & 23.52 & 24.48 & \textbf{24.56} & 23.34 & 23.31 \\
							SSIM $\uparrow$ & 0.917 & 0.919 & \textbf{0.924} & 0.913 & 0.915 \\
							LPIPS $\downarrow$ & 0.172 & 0.153 & \textbf{0.148} & 0.152 & 0.159\\
							FID $\downarrow$ & 8.24 & 7.92 & \textbf{7.70} & 8.04 & 8.16\\
							\bottomrule
						\end{tabular}
				\end{center}}
		\end{minipage}}
		\\
		\centering
		\vspace{-.9em}
	\end{table*}
	\vspace{-0.15in}
	
	\paragraph{PalNorm and Color Discriminator}
	In Tab. \ref{tb_ab_structure}, we find PalNorm yields better quantitative results than BN and SPADE \cite{park2019semantic} (we use gray-input as semantic layout to generate pixel-wise affine transformation). Besides, PalGAN (default with Color Discriminator) beats PalGAN with Patch Discriminator \cite{wang2018high}. These show the effectiveness of our designed PalNorm and Color discriminator.
	
	\vspace{-0.15in}
	\paragraph{Palette Configuration} We systematically explore different factors of the employed palette. Tab. \ref{tb_ab_bins} shows how the number of bins of palette affects the colorization results. Generally, when \#Bins is relatively small, increasing it (16 $\rightarrow$256) will lead to a performance increase on almost all used metrics; while \#Bins is relatively large, increasing it (256 $\rightarrow$1024) will lead to a performance drop. We conjecture this is caused by the tradeoff between the fineness and sparsity of the used palette. The rise of \#Bins enhances both its fineness and sparsity. The former reduces ambiguities of palette and the latter makes the optimization of palette reconstruction harder. Empirically, \#Bins is 256 is an acceptable setting, used in all experiments.
	
	Also, as given in Tab. \ref{tb_ab_structure} (PalGAN wo $E(\mathbf{\hat{h}})$), applying diversity regularization on the estimated palette can improve our generative performance.
	
	\vspace{-0.15in}
	\paragraph{Limitation}
	
	In the user-guided colorization, current PalGAN lacks fine-grained control as we utilize a global palette. In addition, PalGAN cannot well address small-size regions with independent semantics (\eg\ small objects), since the global interaction in CA cannot well represent these areas using semantic embeddings from small-scale feature maps. Failure cases are given in the supp.
	
	\section{Concluding Remarks}
	In this paper, we study multimodal challenges and color bleeding in colorization from a new perspective. We give a new formulation of colorization for multimodal representation. It introduces the palette as an intermediate variable. This leads to a new and workable colorization method by palette estimation and assignment, yielding diverse and controllable colorful outputs. Additionally, we address the color bleeding issue by explicitly studying color affinities using chromatic attention. It not only leverages semantic affinities to coordinate color, but also exploits the correlation between intensity and their corresponding chrome to delineate color details. Our method is experimentally proven effective and surpasses existing state-of-the-arts non-trivially.
	\vspace{-0.15in}
	\paragraph{Acknowledgment}
	This work is partially supported by the Shanghai Committee of Science and Technology (Grant No. 21DZ1100100).

	\clearpage
	
	\bibliographystyle{splncs04}
	\bibliography{egbib}
\end{document}